\documentclass[journal,twoside,web]{ieeecolor}
\usepackage{generic}
\usepackage{cite}
\usepackage{amsmath,amssymb,amsfonts}
\usepackage{algorithmic}
\usepackage{graphicx}
\usepackage{algorithm,algorithmic}
\usepackage{hyperref}
\hypersetup{hidelinks=true}
\usepackage{arydshln}
\usepackage{multirow}
\usepackage{booktabs}
\usepackage{textcomp}
\def\BibTeX{{\rm B\kern-.05em{\sc i\kern-.025em b}\kern-.08em
    T\kern-.1667em\lower.7ex\hbox{E}\kern-.125emX}}
\begin{document}
\title{Memory-based Cross-modal Semantic Alignment Network for Radiology Report Generation}

\author{Yitian Tao,
        Liyan Ma,
       Jing Yu and Han Zhang, \IEEEmembership{Senior Member, IEEE} 
\thanks{This work was supported in part by the National Key R\&D Program of China (No. 2021YFA1003004), in part by the Shanghai Municipal Natural Science Foundation under Grant 21ZR1423300 (Corresponding author: Liyan Ma).}
\thanks{Yitian Tao was with the School of Computer Engineering and Science, Shanghai University, Shanghai, China. He is now with the School of Biomedical Engineering, Shanghaitech University, Shanghai, China (E-mail: 19120184@shu.edu.cn, taoyt2023@shanghaitech.edu.cn).}
\thanks{Han Zhang is with the School of Biomedical Engineering, Shanghaitech University, Shanghai, China (E-mail: zhanghan2@shanghaitech.edu.cn).}
\thanks{Jing Yu is with the Institute of Information Engineering, Chinese Academy of Sciences, Beijing, China (E-mail: yujing02@iie.ac.cn).}
        \thanks{Liyan Ma is with the School of Computer Engineering and Science, Shanghai University, Shanghai, China (E-mail: liyanma@shu.edu.cn).}}




\maketitle

\begin{abstract}
Generating radiology reports automatically reduces the workload of radiologists and helps the diagnoses of specific diseases. Many existing methods take this task as modality transfer process. However, since the key information related to disease accounts for a small proportion in both image and report, it is hard for the model to learn the latent relation between the radiology image and its report, thus failing to generate fluent and accurate radiology reports. To tackle this problem, we propose a memory-based cross-modal semantic alignment model (MCSAM) following an encoder-decoder paradigm. MCSAM includes a well initialized long-term clinical memory bank to learn disease-related representations as well as prior knowledge for different modalities to retrieve and use the retrieved memory to perform feature consolidation. To ensure the semantic consistency of the retrieved cross modal prior knowledge, a cross-modal semantic alignment module (SAM) is proposed. SAM is also able to generate semantic visual feature embeddings which can be added to the decoder and benefits report generation. More importantly, to memorize the state and additional information while generating reports with the decoder, we use learnable memory tokens which can be seen as prompts. Extensive experiments demonstrate the promising performance of our proposed method which generates state-of-the-art performance on the MIMIC-CXR dataset.
\end{abstract}


\section{Introduction}
\label{sec:introduction}
\IEEEPARstart{R}{adiology} reports written by expertized radiologists are always used for diagnoses of specific diseases. However, writing radiology reports is a labour-intensive task which requires rigorous analyses of radiology images, and this makes the task error-prone and time-consuming \cite{6966720,liu-PPKED}. With the development of deep neural networks, as an efficient method for assisting radiologists in writing reports, automatic radiology report generation, which requires the model to generate accurate and fluent reports according to the radiology images, has attracted increasing attention and a lot of efforts have been made.

Inspired by image captioning \cite{caption1, caption2, caption3, caption4, caption5}, encoder-decoder architecture including a CNN-based encoder and an RNN-based decoder is also popular in medical report generation task. However, abnormalities and their corresponding terminologies only exist in small parts of the radiology images and reports, which makes it hard for the model to learn the abnormality representations of both image and text modalities. Thus, automatic report generation is far from satisfactory.

To mitigate this problem, human-assisted processes and different kinds of structured labels are widely used in some previous studies. Some works used a knowledge graph based paradigm, where a pre-defined knowledge graph constructed from chest x-ray radiology reports is leveraged to get correlation of different diseases and embed all the related disease representations into either the visual or textual representations \cite{liu-PPKED, MGSK}. Others used pre-defined disease labels to perform a classification task on the generated report \cite{learning-clinically-coherent, nguyen2021automated}, which directly force the model to generate reports with high factual completeness. However, both knowledge graphs and classification tasks require the radiologists to label the data. If the dataset changes or new types of diseases are required to be classified, the data need to be labeled again, which adds the burden of the radiologists. 


Practically, when a radiologist writes a report according to a radiology image, clinical knowledge (or memory) such as correlation of different abnormalities which can indicate specific diseases is already in his head. After comparing the information extracted from a specific image with memory, a radiologist can quickly diagnose the disease and write the corresponding report. Inspired by this, we firstly formulate a memory bank initialized by learning the topics (e.g., correlation of different abnormalities) of radiology reports, which are disease-related representations shared between different modalities and use a cross attention based method to retrieve the memory bank. Then, via incorporating the retrieved memory which can be seen as prior knowledge into cross modality representations, we can achieve fine-grained feature consolidation. After that, a cross-modal semantic alignment module is proposed to ensure the semantic consistency of the retrieved cross modal prior knowledge and generate semantic visual feature embeddings which can be added to the decoder and benefit report generation. What's more, to generate more fluent sentences, we also use some learnable tokens in the decoder which can be seen as prompts that can store some additional information to benefit the generation performance.

In general, the contributions of our work can be summarized as follows:
\begin{itemize}
    \item We design a carefully initialized memory bank to learn disease-related representations shared between different modalities and use the retrieved memory to perform cross modal feature consolidation. Thus, the model can focus more on abnormalities and alleviate the data bias problem.
    \item A cross-modal alignment module is proposed to ensure the semantic consistency of the retrieved cross modal knowledge and produce a semantic embedding which can be added to the decoder and benefits report generation.
    \item Comparison results show the superiority of our method, even those using structured labels or pre-constructed knowledge graphs, neither of which are required in our method.
\end{itemize}

\section{Related Works}
\subsection{Medical Report Generation.} 
Due to the development of image captioning methods, existing methods are mainly based on an encoder-decoder architecture and adopt different techniques to generate better medical reports. Wang \textit{et al.} \cite{wang2018tienet} developed a CNN-LSTM structure which used the attention mechanism to capture disease related details and generate reports. Miura \textit{et al.} \cite{improving-factual} mainly followed an image captioning model $M^2$ Transformer \cite{cornia2020meshed} and designed two rewards for reinforcement learning to improve factual completeness and consistency of the generated report. Lovelace \textit{et al.} \cite{lovelace2020learning} introduced the clinical information to the report generation process tried to generated more clinical correct reports. Wang \textit{et al.} \cite{medical-semantic} tried to get additional entities as labels from a pre-defined knowledge graph and proposed a medical concepts generation network to get enriched semantic information to promote report generation. Li \textit{et al.} \cite{dynamic-grapg-enhace} leveraged a dynamic knowledge graph to get prior knowledge and enhance the visual representations. Li \textit{et al.} \cite{A-Self-guided} firstly presented a self-guided framework to obtain potential medical knowledge, and then adopted the Sentence Bert to maximize the similarities between ground truth and the generated report. Some of these works still require human-assisted process such as pre-constructing a knowledge graph or producing labels by radiologists. Since focusing on the single modality feature consolidation and paying more attention on report generation, other works fail to understand the inherent relationship between radiology images and their corresponding reports. 

To exploit the inherent correlation and interaction between different modalities, Wang \textit{et al.} \cite{wang2022medclip} and Ramos \textit{et al.} \cite{9894658} utilized image-text pretraining for facilitating downstream tasks. Najdenkoska \textit{et al.} \cite{najdenkoska2021variational} proposed a variational topic inference method which addressed report generation task with a probabilistic latent variable model and optimized the model by maximizing an evidence lower bound objective (ELBO). Wang \textit{et al.} \cite{self-boosting} adopted a self-boosting framework which uses a hierarchical LSTM to generate reports and enhance the model performance with an auxiliary image-text matching task. Chen \textit{et al.} \cite{chen2020generating} utilized a gate mechanism in the transformer decoder to help with the report generation, Chen \textit{et al.} \cite{chen2022cross} and Wang \textit{et al.} \cite{cross-proto} adopted either a randomly initialized parameter matrix or a prototype based representation matrix to formulate the unknown intermediate representations. However, neither of them proposed an interpretable or effective way to initialize the parameter matrix and make constraint on the model. Thus, the training of those models is unstable. 

\subsection{Optimal Transport}
Optimal Transport(OT)\cite{monge1781memoire} distance is a key method for comparing probability measures. It has been well adopted to many applications such as Neural Topic Model (NTM)\cite{zhao2020neural-topic-model},  or evaluation of textual similarity\cite{lee2022ot_for_text, xu-etal-2021-ot-vocabulary}. The  OT distance $d_P$ is defined for discrete distributions  $r \in \mathbb{R} ^{D_r}$,  $c \in \mathbb{R} ^{D_c}$ as the smallest Frobenius dot-product between a transport matrix $A \in \mathbb{R}_{>0} ^{D_r \times D_c}$ and a cost matrix $B \in \mathbb{R}_{\ge0} ^{D_r \times D_c}$:
\begin{equation}
    d_B(r, c)=\min_{A\in U(r, c)} <A, B>,
\end{equation}
where $U(r, c)=\{A\in\mathbb{R}_{>0} ^{D_r \times D_c}|A1_{D_c}=r, A^T_{1_{D_r}}=c\}$ is the  polyhedral set which denotes the transportation polytope of $r$ and $c$. The transport matrix links two discrete probability distributions in a way that represents their joint probabilities. The optimal transport matrix is calculated to find the optimal transport distance.

However, this calculation can be cumbersome for large-scale cases. Thus, a streamlined version called the Sinkhorn distance\cite{cuturi2013sinkhorn} which includes an entropy-based regularization term in its formulation is employed:
\begin{equation}
    d_{B,\alpha}(r, c)=\min_{A\in U_{\alpha}(r, c)} <A, B>,
\end{equation}

\begin{equation}
    U_{\alpha}(r, c)=\{A\in U(r, c)|h(A)\ge h(r)+h(c)-\alpha\}.
\end{equation}

The Sinkhorn distance uses a Lagrange multiplier to solve for the minimized regularized optimal transport distance, making it a practical solution for discrete transport issues.

\section{Methodology}
\begin{figure*}[h]
  \centering
  \includegraphics[width=1\linewidth]{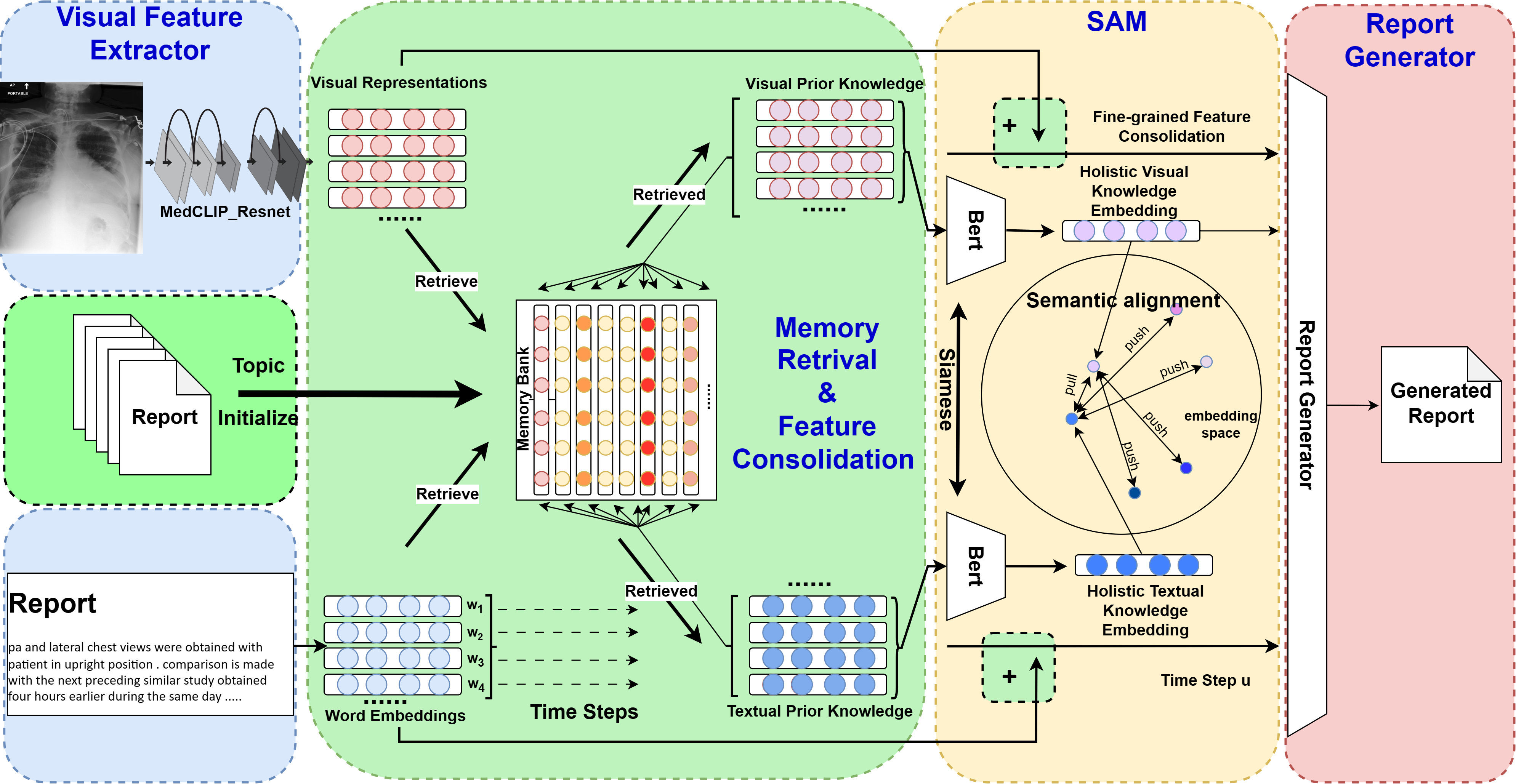}
  \caption{Illustration of MCSAM, which can be divided into three parts: a cross modal memory bank, a cross modal semantic alignment module (SAM) and a report generator. Cross modal memory bank is used to learn disease-related representations as well as prior knowledge for different modalities to retrieve and use the retrieved memory to perform feature consolidation. Cross modal semantic alignment module (SAM) is proposed to ensure the semantic consistency of the retrieved cross modal prior knowledge. The report generator generates reports depending on the retrieved memory and learnable prompts.}
  \label{figure:1}
\end{figure*}

\subsection{Overview}
Radiology report generation task requires the model to generate radiology reports according to their corresponding radiology images, and the data can be seen as the composition of large amounts of (image, text) pairs. To generate accurate and fluent reports, our proposed model can be divided into three parts: a Image feature Extractor, a Cross modal Memory bank, a Cross modal Semantic Alignment Module(SAM) and a Report Generator. 

Before training the whole model, inspired by brain-like memory, we use optimal transport algorithm to initialize the memory bank $M$ via learning topics (e.g., correlation of different abnormalities) of radiology reports which act as disease-related representations (memory of clinical knowledge). 


At the training stage, given a radiology image $I\in \mathbb{R}^{H\times W \times C}$, a visual feature extractor Resnet50 \cite{resnet} $f(.)$ pretrained by MedCLIP \cite{wang2022medclip} is used for image feature extraction: $V=f(I) \in\mathbb{R}^{N_I \times D}$, where $N_I$ is the number of visual representation vectors and $D$ is the feature dimension. At the same time, the corresponding report of the radiology image is translated into word embeddings. After that, both visual representations and word embeddings (textual representations) are used for memory retrieval process for feature consolidation, the consolidated feature representations are leveraged to generate reports by the report generator and the retrieved memory which is called prior knowledge will be send to SAM to do semantic alignment. The radiology report generation task can be formulated as :
\begin{equation}
    p(T|I)=\prod_{t=1}^l p(w_t|w_1,w_2,...,w_{t-1},f(I), M),
\end{equation}
where $T=\{w_1, w_2, w_3....w_l\}$ is the report, $w_i$ is the $i^{th}$ word in the report and $l$ is the report length. Our proposed model is then trained to maximize $p(T|I)$.

\begin{figure}[h]
  \centering
  \includegraphics[width=0.9\linewidth]{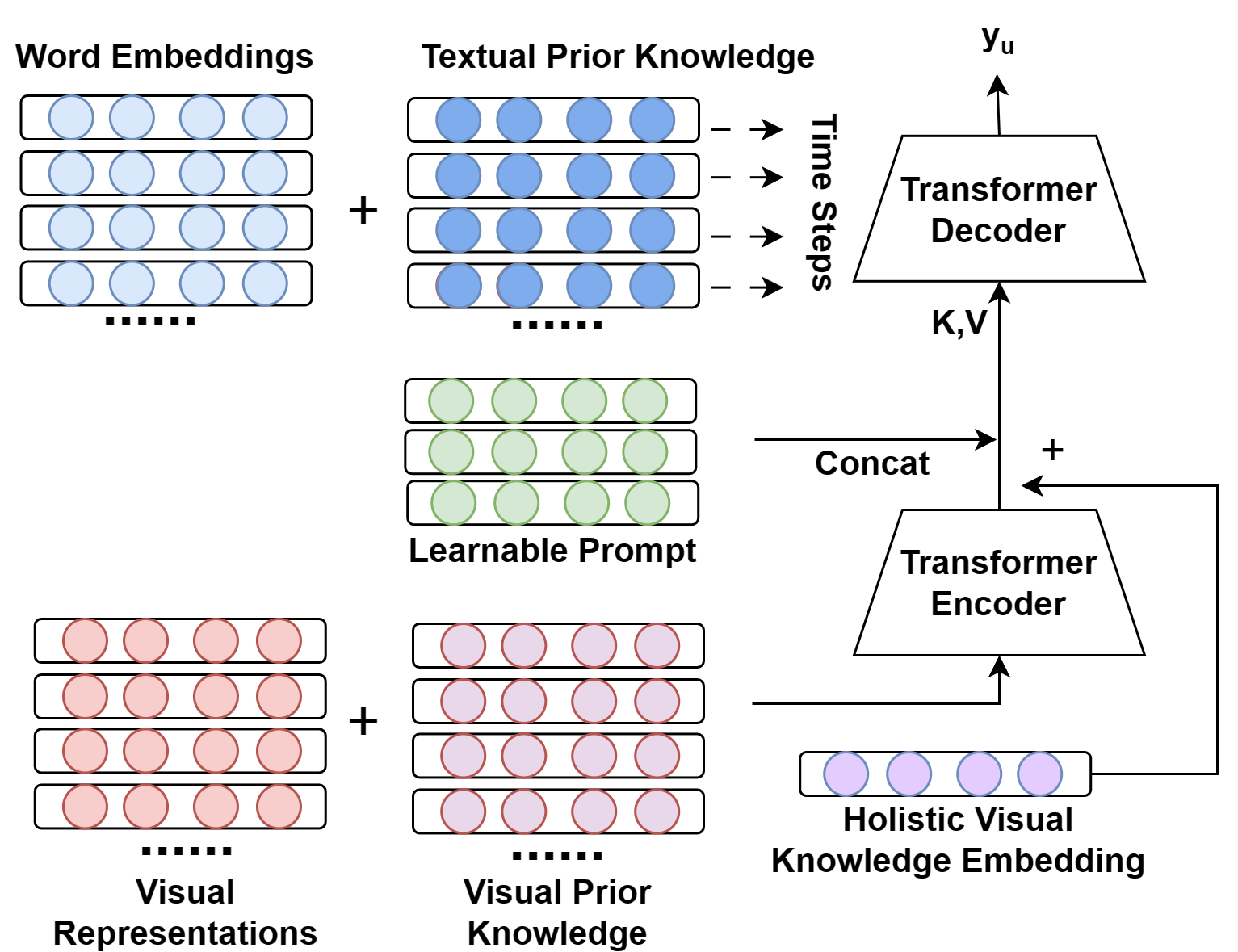}
  \caption{The report generator consists of a visual encoder and a report decoder with learnable prompts. }
  \label{figure:2}
\end{figure}

The whole model structure is showed in Figure \ref{figure:1}, the report decoder structure is showed in Figure \ref{figure:2} and the details of our method are described in following sections.

\subsection{Cross Modal Memory Bank}
\textbf{Memory Bank Construction.} To exploit the inherent relationship between radiology images and their corresponding reports more directly and make our model more interpretable, we propose a carefully designed memory bank $M\in{\mathbb{R}}^{N^m \times D_M}$ ($N^m$ is the capacity of the memory bank and $D_M$ is the feature dimension) which acts as a kind of long term memory for disease-related representations shared between different modalities.

If we randomly initialize our memory bank \cite{chen2022cross}, problems would arise. Firstly, due to the low constraint on the memory bank, more effort (e.g., more training time) will be made to train the model. Secondly, model performance varies considerably because of the randomness of initialization. According to the dual-coding knowledge neural representation framework \cite{bi2021dual, wang2020two} and its corresponding experiments, some knowledge such as relations of symbols is language-derived and it can not be derived easily from other sensory such as vision. Additionally, there is natural high correlation between regions related to sensory-derived knowledge and language-derived knowledge in human brain, and sometimes without the sensory-based knowledge (such as vision), specific tasks can still be accomplished, which indicates that adequate and key information can be obtained from the radiology reports and we can store that to initialize the cross-modal memory bank. To achieve this, inspired by the neural topic model \cite{zhao2020neural-topic-model}, we try to learn topics from the reports in the dataset by using optimal transport algorithm and use the learned topic embeddings to initialize our memory bank.

Specifically, for a single report $T=\{w_1, w_2, w_3....w_l\}$, we firstly calculate the occurrence  $t\in \mathbb{R}^{V_d}$ of each word ( $t_i$ is the occurrence number of $i^{th}$ word in the vocabulary. If the $i^{th}$ word doesn't appear in the report then $t_i$ becomes zero). We translate $t$ into a distribution over $V_d$ words $\tilde{t}\in \Delta^{V_d}$ by normalizing $t$ with $l$. In addition, each report is associated with $N^m$ topics whose distribution is $z\in\Delta^{N^m}$, and $z_j$ denotes the proportion of topic $j$ in the report. In practice, topic distribution $z$ is generated from $\tilde{t}$ using a multi-layer perceptron (MLP) with dropout layers to add randomness:  $ z=softmax(MLP(\tilde{t}))$. Then we define a cost Matrix $P\in\mathbb{R}^{V_d \times N^m}$ for the optimal transport (OT) algorithm :
\begin{equation}
    P_{ij} = 1-cos(e_i, g_j),
\end{equation}
where $cos(.\ , .)$ is the cosine similarity, $e_i\in\mathbb{R}^{D_M}$ and $g_j\in\mathbb{R}^{D_M}$ are the embeddings of the $i^{th}$ word and the $j^{th}$ topic, respectively. We denote  $E\in\mathbb{R}^{V_d \times D_M}$ as the collection of word embeddings $e$ which is initialized with the word embeddings of a pretrained MedCLIP \cite{wang2022medclip}. 

To learn the topic embeddings $G\in\mathbb{R}^{N^m\times D_M}$ which is randomly initialized, we follow \cite{zhao2020neural-topic-model} and use a decoder: $\phi(z):=softmax((2-P)z)$ to induce the loss function for memory bank pre-training (only for pretraining):
\begin{equation}
    \min_{\theta, G}(-\varepsilon\tilde{t}^T log\phi(z)+d_{P,\alpha}(\tilde t, z)),
\end{equation}
where $\varepsilon$ is a hyperparameter which controls the loss contribution and $\theta$ is the model parameters of the MLP. The first part of the joint loss is the expected multinomial log-likelihood. It is proved in 
 \cite{frogner2015learning-with-wasser} that combining the OT loss with the expected multinomial log-likelihood helps learn better topics and  brings better model performance. The second part is a Sinkhorn distance, where $\alpha$ is its hyperparameter. To compute the loss, Sinkhorn algorithm \cite{cuturi2013sinkhorn} is performed and it won't stop until convergence. By optimizing the objective, each report can be related to its own topics with the shortest OT (optimal transport) distance and the topic embeddings can be learned during this process. What's more, the word embeddings used in the OT algorithm come from the multi-modality contrastive learning pretrained model MedCLIP, which helps mitigate the gap between different modalities (or domain shift) and thus contribute to the subsequent whole model training period.

Different from directly clustering and averaging the representations \cite{cross-proto} coming from a Resnet or ViT \cite{ViT} model, our method is more interpretable and shows better performance in the experiments.

\textbf{Memory Retrieval and Feature Consolidation}. Since the memory bank is carefully initialized, its crucial to fully utilize the memory bank and incorporate it into the report generation process. Practically, we use a cross attention paradigm to retrieve useful representations which can be seen as prior knowledge from the memory bank and add them to representations of image and text modalities to make feature consolidation.

For a single radiology image, visual representations $V=\{v_1,v_2,...,v_{N_I}\}=f(I)$ are produced. As for its corresponding report of length $l$, we use the word embeddings: $R={\{\textbf{w}_1, \textbf{w}_2, \textbf{w}_3..., \textbf{w}_l \}}$. Before memory retrieving, linear transformation is performed on the memory bank by a matrix $W^L\in\mathbb{R}^{D_M\times D}$  to generate a memory matrix which has the same dimension $D$ as visual and textual representations: $ \mathbf{M}=MW^L$. Then, additional linear transformation matrices $W^Q$, $W^K$ and softmax functions are employed to get similarity maps between single-modality representations and memory matrix:
\begin{equation}
    Q^V=VW^Q,Q^R=RW^Q,K=\mathbf{M}W^K,
\end{equation}
\begin{equation}
    J^V = Softmax(\frac{Q^VK}{\sqrt{D}}), J^R=Softmax(\frac{Q^RK}{\sqrt{D}}),
\end{equation}
where $J^V$ and $J^R$ are image-memory and report-memory similarity maps, respectively.

Then, following \cite{chen2022cross}, for a single modality representation vector $v_i$ or $w_i$ , we choose top k elements in the memory matrix that have the highest similarity scores with the representation vector. Simply calculating the weighted average with a softmax function performed on similarity scores, the final prior knowledge $P^V$ and $P^S$ are retrieved from the memory bank:
\begin{equation}
    P^V=Softmax(C^V_{sim})(\mathbf{M}^V_{retrieved}W^V),
\end{equation}
\begin{equation}
    P^R=Softmax(C^R_{sim})(\mathbf{M^R}_{retrieved}W^V),
\end{equation}
where $W^V$ is adopted as a linear transformation matrix for the elements selected from the memory matrix, and $\mathbf{M}^V_{retrieved}$, $\mathbf{M}^R_{retrieved}$,  $C^V_{sim}$, $C^R_{sim}$ are top k elements and similarity scores for visual and textual representations, respectively.

To incorporate the retrieved prior knowledge into report generation process, we simply add it to its corresponding modality representations to get the fine-grained feature consolidation:
\begin{equation}
    V' = V + P^V, R' = R + P^R.
\end{equation}

\subsection{Cross Modal Semantic Alignment Module (SAM)}
Though the memory bank is already carefully initialized, challenges still exist during training since there is no more direct constraint on the memory bank optimization and the memory retrieval process. Intuitively, the semantic information of retrieved prior knowledge $P^R$ and $P^V$ derived from a specific (image, text) pair should be the same. Thus, we introduce a cross-modal semantic alignment module (SAM) which is inspired by contrastive learning to constrain the memory retrieval process.

For visual prior knowledge and textual prior-knowledge retrieved from the memory bank, we use a siamese structure network consists of Bert \cite{devlin2018bert} to translate them into [CLS] tokens $S^R$, $S^V\in\mathbb{R}^D$ which serve as semantic feature embeddings of different modalities :
\begin{equation}
    S^R=Bert(P^R)_{[cls]}, \quad S^V=Bert(P^V)_{[cls]}.
\end{equation}
Then, for a single batch of (image, text) pairs which are translated into semantic feature embeddings $\{(S^V_1,S^R_1)_1, (S^V_2,S^R_2)_2,..,(S^V_{N_{batch}},S^R_{N_{batch}})_{N_{batch}}\}$, we firstly define a similarity score $sim_{ij}=cos(S^V_i, S^R_j)$ and then adopt the contrastive loss to push the semantic feature embedding from the same pair(but different modalities) closer and others apart:


\begin{equation}
\label{eq:Positional Encoding}
\begin{split}
    L_{align}=-\frac{1}{N_{batch}}(\sum_{i=1}^{N_{batch}} log(\frac{exp \ sim_{ii}/ \tau}{\sum_{j=1}^{N_{batch}}exp  \ sim_{ij}/ \tau}) \\
    +\sum_{j=1}^{N_{batch}} log(\frac{exp \ sim_{jj}/ \tau}{\sum_{i=1}^{N_{batch}}exp  \ sim_{ij}/ \tau})),
\end{split}
\end{equation}
where $\tau$ is a temperature parameter and $N_{batch}$ is the batch size.

The use of the siamese Bert can be referred to as amortized optimization \cite{marino2018iterative} and has twofold benefits. Firstly, since the model itself has to fit the input coming from different modalities, it can reach an optimal point easier and reduce the training loss more efficiently. Thus, the cross modality task produces natural constraint on the siamese network. Secondly, to reduce the loss, the retrieved prior knowledge of different modalities can be more easily aligned when the same transformation is applied to them. In addition, we also make comparison between different alignment methods, which is included in section IV.E (The Effect of Different Alignment Methods).

\subsection{Report Generator}
\textbf{Visual Encoder.} After the semantic alignment process which is attained by siamese Bert, the semantic feature embedding of visual modality is also incorporated into report generation. A transformer encoder $f_{encoder}(.)$ is adopted to $V'$ to get the final visual representations: $V''=f_{encoder}(V')$. Then, the semantic visual feature embedding $S^V$ is added to $V''$, and we obtain:
\begin{equation}
    \mathbf{V} = V''+S^V.
\end{equation}

The application of this integration is almost the same as the feature consolidation process but have different meanings. Firstly, the semantic visual feature embedding, which is actually learned by cross-modal alignment, may contain some mutual information shared between different modalities. Thus, it can be interpreted as the main information of an image and its corresponding reports, and benefit the report generation task. Secondly, the integration of $S^V$ and $V''$, which denotes that extra participation of memory bank is added to the workflow of our model, can facilitate the learning of an excellent memory bank.

\textbf{Report Decoder with Learnable Prompts.} Prompts are used in large language models such as gpt to help with specific downstream tasks, some previous works\cite{cornia2020meshed}\cite{li2023efficient}\cite{bulatov2022recurrent} also combine additional learnable parameters with transformer models. Inspired by this, we develop our report decoder which have additional learnable parameters to serve as prompts.

Traditionally, to generate a single word $y_u$  with transformer decoder at time step $u$, the output $\mathbf{V}$  of the visual encoder and consolidated word embeddings of previous time steps are fed into the decoder $f_{decoder}$ (.):
\begin{equation}
    y_u=f_{decoder}(\mathbf{V}, R_1',R_2',....R_{u-1}').
\end{equation}

We combine $\mathbf{V}$  with some randomly initialized parameters $H=\{h_1,h_2..h_{N_h} \} \in\mathbb{R}^{N_h\times D}$ which act as learnable prompts and try to memorize some states and additional information that are also critical to generate a report:
\begin{equation}
    y_u=f_{decoder}([\mathbf{V},H], R_1',R_2',....R_{u-1}').
\end{equation}
Then, the report generation loss can be formulated as a cross entropy loss:
\begin{equation}
    L_{gen}=-\frac{1}{l}\sum_{i=1}^l \sum_{j=1}^{V_d}w_{ij}log(y_{ij}),
\end{equation}
where $l$ is the length of the report, $V_d$ is the vocabulary size, $w_{ij}$  and $y_{ij}$ are the $j_{th}$ element of ground truth one-hot vector of the $i_{th}$ word and the predicted word, respectively. 

Collectively, our model is trained by minimizing the joint loss consisting of both report generation loss and semantic alignment contrastive loss :

\begin{equation}
    L=L_{gen}+L_{align}.
\end{equation}

\begin{table*}
\vspace{-0.7cm}
  \caption{Comparison with the State-of-the-Arts. \dag indicates the results are quoted from the published paper while \ddag indicates the results are obtained by the publicly released codes}
  \centering
  \label{table:1}
  \begin{tabular}{c|c|cccccc|c}
    \toprule
    Dataset & Method & BLEU-1 & BLEU-2 & BLEU-3 & BLEU-4 & METEOR & ROUGE-L & Human-assisted\\
    \midrule
    \multirow{11}{*}{MIMIC-CXR} & R2Gen\dag & 0.353 &0.218 &0.145 &0.103 &\underline{0.142} &0.277 & No\\
    & ATT2IN\dag & 0.325 & 0.203 & 0.136 & 0.096 & 0.134 & 0.276 & No \\
    & PPKED\dag & 0.360 & 0.224 & 0.149 & 0.106 & \textbf{0.149} &\textbf{ 0.284} & Yes\\
    & R2GenCMN\ddag & 0.344 & 0.210 & 0.141 & 0.100 & 0.138 & 0.271 & No\\
    & CMCL\dag & 0.344 & 0.217 & 0.140 & 0.097 & 0.133 & \underline{0.281} & No\\
    & Self-boost\dag & 0.359 & 0.224 & 0.150 & 0.109 & 0.141 & 0.277 & No\\
    & MGSK\dag & 0.363 & 0.228 & \underline{0.156} & \underline{0.115} & / & \textbf{0.284} & Yes \\
    & XPRONET\ddag & 0.344 & 0.215 & 0.146 & 0.105 & 0.138 & 0.279 & Yes\\
    & PGT\dag & 0.356 & 0.222 & 0.151 & 0.111 & 0.140 & 0.280 & No\\
    & ITA\dag &\textbf{0.395} & \textbf{0.253} & \textbf{0.170} & \textbf{0.121} & 0.147 & \textbf{0.284} & Yes\\
    & MCSAM (Ours) & \underline{0.379} & \underline{0.230} & 0.153 & 0.109 & \textbf{0.149} & \textbf{0.284} & No \\
    \hdashline
    & CvT-DistilGPT2\dag & 0.394 & 0.249 & 0.172 & 0.127 & 0.155 & 0.287 & No\\
    & RGRG\dag & 0.373 & 0.249 & 0.175 & 0.126 & 0.168 & 0.264 & Yes\\
    \midrule
    \multirow{7}{*}{IU-Xray} & ADAATT\dag & 0.299 &0.185 &0.124 &0.088 &0.118 &0.266 & No\\ 
    & CMAS\dag &0.464 & 0.301 & 0.210 & 0.154 & / & 0.362& No\\
    & COATT\dag & 0.455 & 0.288 & 0.205 & 0.154  & / & 0.369 & Yes\\
    & R2Gen\dag &0.470 & 0.304 & 0.219 & 0.165 & 0.187 & 0.371 & No\\
    & CMCL\dag & 0.473 & 0.305 & 0.217 & 0.162 & 0.186 & 0.378 & No\\
    & R2GenCMN\ddag &0.474 & 0.302 & 0.220 & 0.168 & 0.198 &0.370 & No\\
    & PPKED\dag & 0.483 & 0.315 & 0.224 & 0.168 & / & 0.376 & Yes\\
    & JPG\dag &0.479 & 0.319 &0.222 &0.174 &0.193 &0.377 & No\\
    & PGT\dag & 0.482 & 0.313 & 0.232 & 0.181 & 0.203 & 0.381 & No \\ 
    & ITA\dag & \textbf{0.505} & \textbf{0.340} & \textbf{0.247} & \textbf{0.188} & \underline{0.208} & \underline{0.382} & Yes \\ 
    & MCSAM(Ours) & \underline{0.489} & \underline{0.325} & \underline{0.238} & \underline{0.184} & \textbf{0.210} & \textbf{0.394} & No\\
    \hdashline
    & CvT-DistilGPT2\dag & 0.477 & 0.308 & 0.227 & 0.177 & 0.203 & 0.377 & No\\
    \bottomrule
  \end{tabular}
\end{table*}

\begin{figure*}
  \vspace{-0.7cm}
  \centering
  \includegraphics[width=\linewidth]{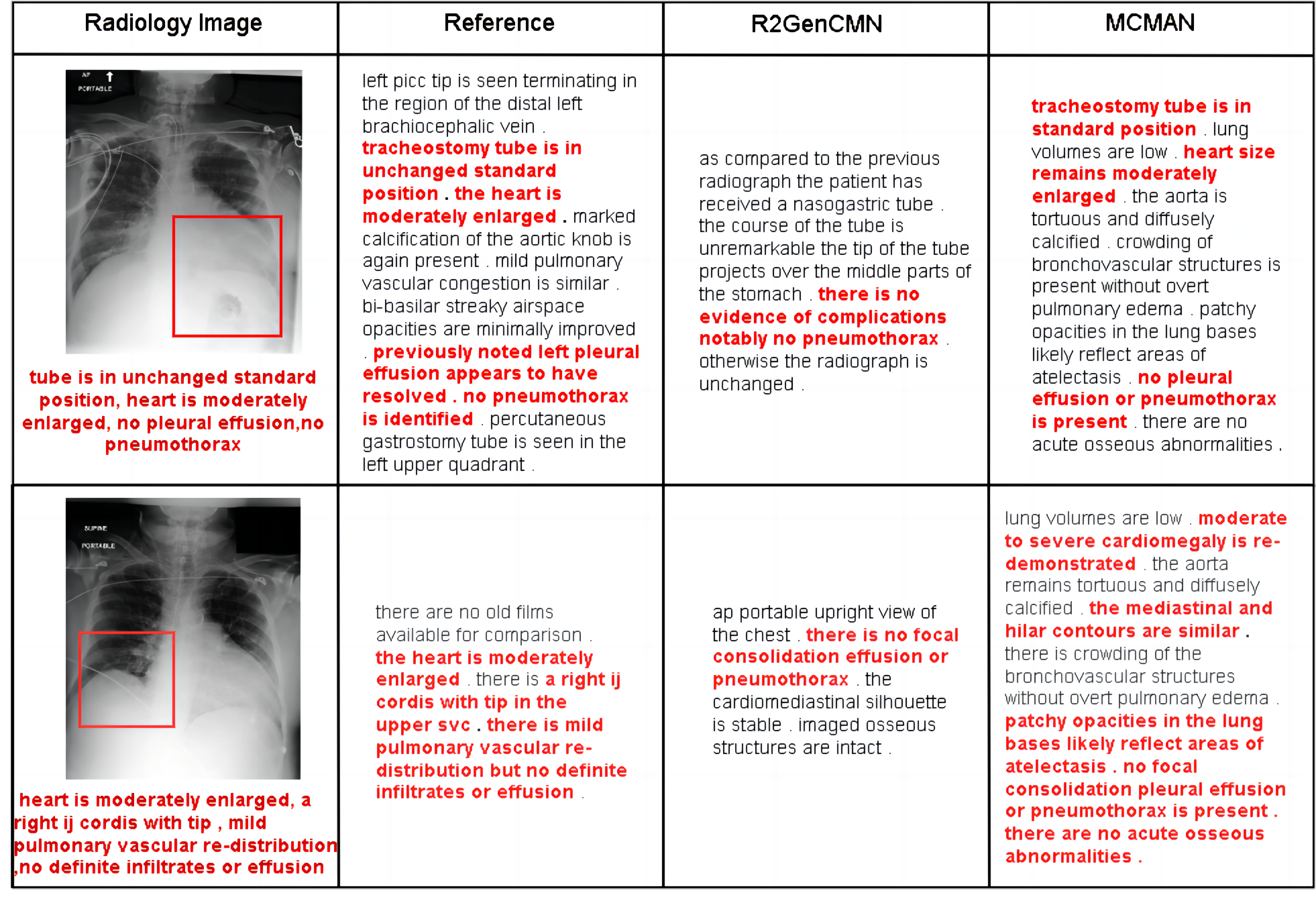}
  \caption{An example to illustrate the quality of the report generated by our model. It can be seen that our model can generate fluent report and have high factual completeness. For example, our model correctly describes "tracheostomy tube is in standard position ", "heart size remains moderately enlarged" and "no pleural effusion or pneumothorax is present".}
  \label{figure:3}
\end{figure*}

\begin{table*}[h]
  \caption{Ablation studies on different modules proposed in our method. The BASE model is conducted with the same encoder-decoder backbone in our whole model, a randomly initialized memory bank, and a pretrained visual feature extractor which comes from a MedCLIP model. \textbf{PreMem}, \textbf{wm} and \textbf{SAM} denote the use of the memory bank initialization trick, the learnable prompts and the cross model semantic alignment network, respectively. All the models are trained and tested on MIMIC-CXR dataset.}
  \label{table:2}
  \centering
  \begin{tabular}{cccccccc}
    \toprule
    Trial & Method & BLEU-1 & BLEU-2 & BLEU-3 & BLEU-4 & METEOR & ROUGE-L \\
    \midrule
    1 & BASE & 0.347 & 0.210 & 0.138 & 0.097 & 0.138 & 0.272 \\
    2 & BASE + PreMem & 0.356 & 0.218 & 0.144 & 0.102 & 0.143 & 0.276 \\
    3 & BASE + PreMem + wm & 0.359 & 0.217 & 0.145 & 0.103 & 0.143 & 0.278 \\
    4 & BASE + SAM + wm & 0.366 & 0.222 & 0.147 & 0.104 & 0.144 & 0.275\\
    5 & BASE + PreMem + SAM & 0.376 & 0.229 & 0.152 & 0.108 & 0.147 & 0.279\\
    6 & BASE + PreMem + SAM + wm & \textbf{0.379} & \textbf{0.230} & \textbf{0.153} & \textbf{0.109} & \textbf{0.149} & \textbf{0.284} \\
    \bottomrule

  \end{tabular}
\end{table*}

\section{Experiments}
\subsection{Datasets and Evaluation Metrics}
Our experiments are conducted on MIMIC-CXR \cite{johnson2019mimic} and IU-Xray \cite{demner2016preparing}, where the former is the largest dataset consisting of 473,057 chest X-ray images and 227,835 radiology reports, and the latter comprises 7,470 chest X-ray images along with 3,955 reports. For a fair comparison, we adopt MIMIC-CXR's split following \cite{chen2022cross, cross-proto, MGSK}, where 222.8k samples for training, 1.8k samples for validation and 3.3k samples for test. As for the IU-Xray dataset, we follow \cite{chen2022cross, chen2020generating} and use 70\%-10\%-20\% training-validation-testing splits. In our experiments, we convert all the tokens to lowercase characters and remove the words with occurrence frequency under 17.

To evaluate the generated report, following previous works \cite{cross-proto, chen2022cross, liu2022competence, self-boosting, liu-PPKED}, we utilize the standard evaluation protocol and choose the most widely used natural language generation metrics which are BLEU scores (i.e., BLEU-1, BLEU-2, BLEU-3, and BLEU-4) \cite{papineni2002bleu}, ROUGE-L \cite{lin2004rouge} and METEOR \cite{denkowski2011meteor}. 

\subsection{Implementation Details}
In our method, We firstly initialize our memory bank and then train the whole model. 

During memory initialization phase, the memory bank is firstly pretrained by optimal transport algorithm on 222.8k radiology reports on the training split of the dataset. The hyperparameter $\alpha$ is set to 20, $\epsilon$ is set to 0.07, learning rate is set to 1e-4 and $N^m$ is set to 2048.

After memory initialization, the whole model is then trained by all the (image, text) pairs in the training split. The number of both visual encoder transformer layers and report decoder transformer layers is set to 3, and the attention head numbers are 8. The layer number of siamese Bert encoder in cross model semantic alignment module is set to 1. The hyperparameter $\tau$ in contrastive loss is set to 1.0. The optimizer we use is Adam \cite{kingma2014adam}  and we train our model on an A100 with the mini-batch size set to 16, number of epoch set to 50, the learning rates of visual feature extractor and the rest part in our model set to 1e-4 and 5e-4, respectively. At the inference stage, the beam search strategy is adopted and the beam size is 3.
\subsection{Comparison with State-of-the-Art}
Table \ref{table:1} shows the performance comparison on both MIMIC-CXR \cite{johnson2019mimic} and IU-Xray \cite{demner2016preparing} datasets between our method and previous state-of-the-art methods, including CNN-RNN based model (i.e., ADAATT \cite{lu2017knowing}, COATT \cite{jing-etal-2018-automatic}, ATT2IN \cite{ATT2IN}, CMAS \cite{CMAS}), CNN-Transformer models (i.e., R2Gen  \cite{chen2020generating}, Self-boost \cite{self-boosting}, JPG \cite{you-etal-2022-jpg}, PGT \cite{PGT}), and methods using techniques such as curriculum learning (i.e., CMCL \cite{liu2022competence}), knowledge graph (i.e., PPKED \cite{liu-PPKED}, MGSK \cite{MGSK} and ITA \cite{wang2022inclusive}) and memory or prototype based module (i.e., XPRONET \cite{cross-proto}, R2GenCMN \cite{chen2022cross}). The table also includes some other SOTA methods such as CvT-DistilGPT2\cite{nicolson2023improving} and RGRG\cite{tanida2023interactive} for reference (both methods are not strictly comparable with our methods due to their experimental settings). RGRG uses another dataset Chest ImaGenome v1.0.0, which is derived from the MIMIC-CXR dataset but has more annotations such as bounding box coordinates for 29 unique anatomical regions in the chest. As for CvT-DistilGPT2, it contains a pretrained GPT model which has large parameter size while our method only uses a 3-layer transformer. According to Table \ref{table:1}, We can learn that our method outperforms most of the existing methods showed in the table with a large margin, especially those using self-supervised learning method (i.e., Self-boost). We also notice that although some previous works that include human-assisted processes such as pre-defined knowledge graphs (i.e., MGSK, PPKED and ITA) or labels generated by radiologists (i.e., XPRONET), our method still shows comparable results and surpasses some of them (even the GPT based model), especially on ROUGE-L and METEOR metrics, which may indicate that our model can actually generate fluent sentences which are similar to the ground truth report although some words may be different. Besides, the superiority of MCSAM against memory or prototype-based methods (i.e., R2GenCMN and XPRONET) shows the necessity of using both cross modal semantic alignment and a good memory bank initialization strategy.   

\begin{figure*}
  \centering
  \includegraphics[width=\linewidth]{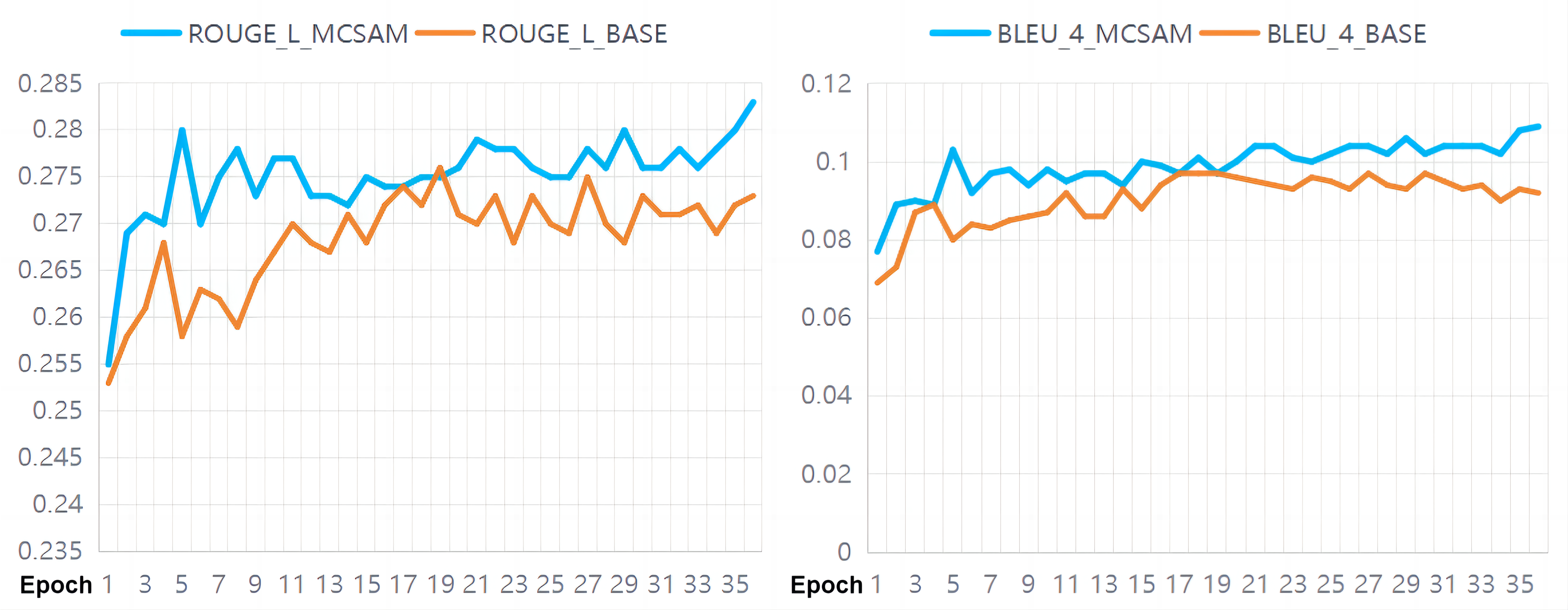}
  \caption{The ROUGE-L and BLEU-4 scores during training, the blue line and the red line denote the performance scores of our method (with the help of OT-initialized memory bank) and baseline method during training, respectively. }
  \label{figure:4}
\end{figure*}

\begin{figure}
  \centering
  \includegraphics[width=\linewidth]{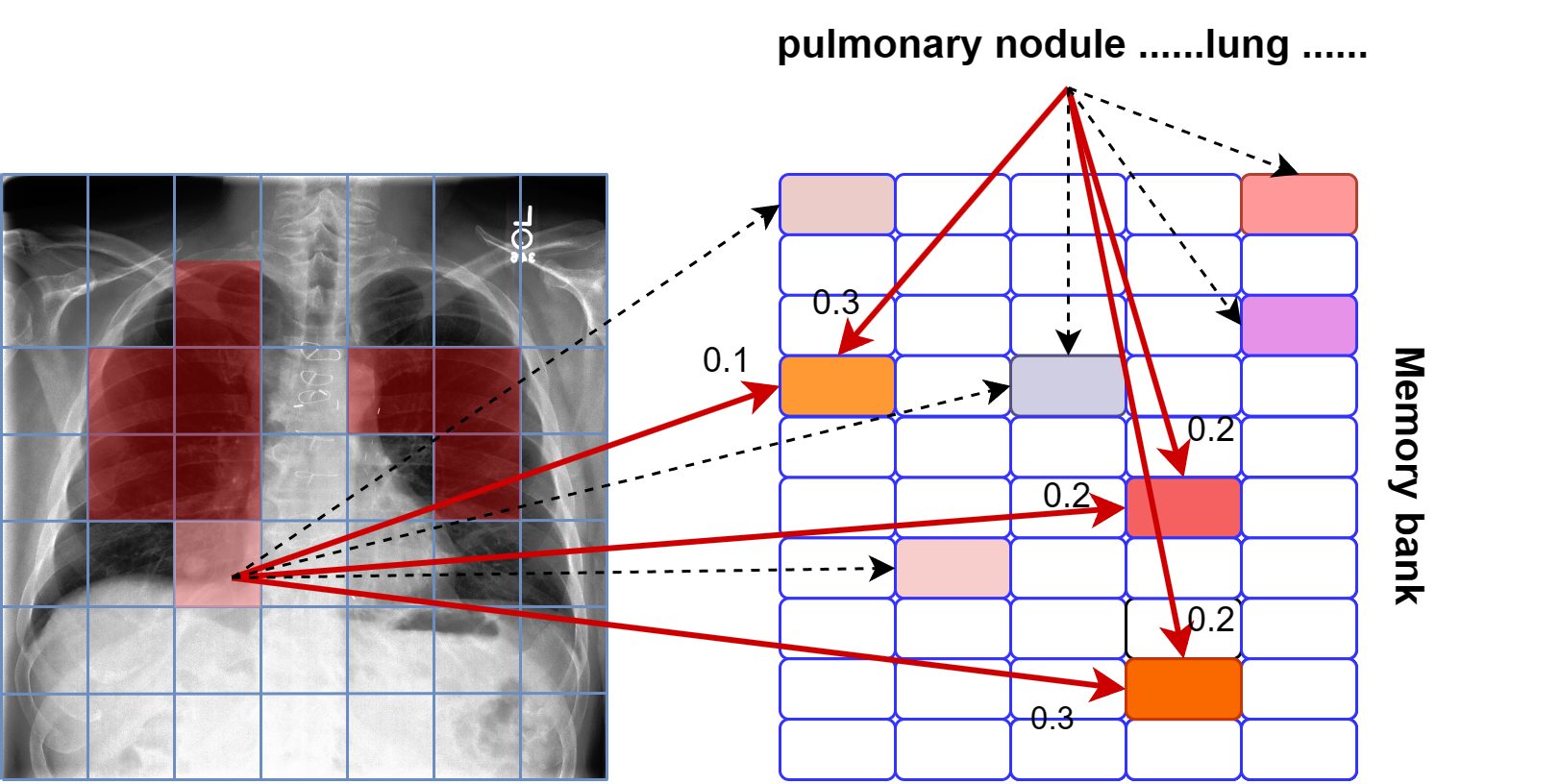}
  \caption{The visualization of memory retrieval process, where the blocks represent the memory vectors in the memory bank, the numbers denote the contribution of specific memory vectors to the construction of prior knowledge, and the red arrows and the black dotted arrows denote the retrieval process of MCSAM and BASE, respectively.}
  \label{figure:5}
\end{figure}

\begin{figure*}
  \centering
  \includegraphics[width=\linewidth]{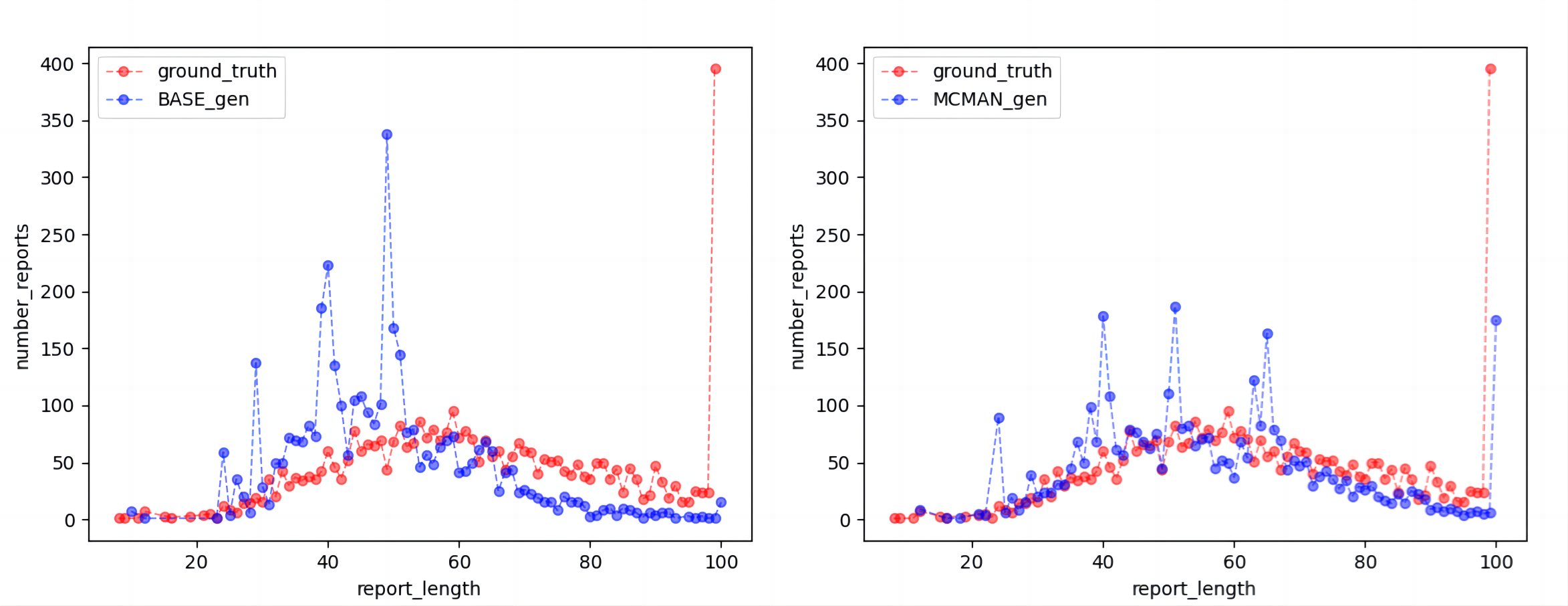}
  \caption{The length distributions of reports generated by BASE and MCSAM. The experiments are conducted on MIMIC-CXR dataset.}
  \label{figure:6}
\end{figure*}

\begin{table}
  \caption{The Effect of Memory Bank Capacity, we conduct the experiments with memory bank capacity of 512, 1024, 2048, 3072 and 4096 on both MIMIC-CXR and IU-Xray dataset.}
  \label{table:3}
  \centering
  \begin{tabular}{ccccc}
    \toprule
     Dataset & $N^m$ & BLEU-4 & METEOR & ROUGE-L \\
    \midrule
    \multirow{5}{*}{MIMIC-CXR} & 512 & 0.098 & 0.138 & 0.273\\
    & 1024 & 0.103 & 0.143 & 0.277 \\
    & 2048 & \textbf{0.109} & \textbf{0.149} & \textbf{0.284}\\
    & 3072 & 0.106 & 0.145 & 0.278\\
    & 4096 & 0.105 & 0.144 & 0.275\\
    \midrule
    \multirow{5}{*}{IU-Xray} & 512 & 0.137 & 0.161 & 0.338\\
    & 1024 & 0.171 & 0.206 & 0.361 \\
    & 2048 & \textbf{0.184} & \textbf{0.210} & \textbf{0.394}\\
    & 3072 & 0.147 & 0.181 & 0.356\\
    & 4096 & 0.133 & 0.160 & 0.349\\
    \bottomrule
  \end{tabular}
\end{table}

\begin{table}
  \caption{The Effect of $k$ (the experiments are conducted on MIMIC-CXR dataset).}
  \label{table:4}
  \centering
  \begin{tabular}{ccccc}
    \toprule
      $k$ & BLEU-4 & METEOR & ROUGE-L \\
    \midrule
    8 & 0.097 & 0.136 & 0.273\\
    16 & 0.104 & 0.144 & 0.276 \\
    32 & \textbf{0.109} & \textbf{0.149} & \textbf{0.284}\\
    64 & 0.102 & 0.143 & 0.274\\
    128 & 0.101 & 0.143 & 0.272\\
    256 & 0.099 & 0.139 & 0.272\\
    \bottomrule
  \end{tabular}
\end{table}

\subsection{Ablation Study}
In this part, we try to analyse some of the implementation details of our model structure and demonstrate the effect of different modules. 

Figure \ref{figure:3} shows the generated report produced by our proposed model, where sentences that include important information are highlighted by radiologists. We can learn that in in both two examples, our model can generate accurate and fluent report. Notably, in the second example, the report generated by our MCSAM model is longer than the reference report(ground truth) and include more details in the radiology image, which imply that the model accurately learn the characters of specific diseases and store them in the memory bank.

According to Trial 1, 2, 4, 6 in Table \ref{table:2} that with the use of a pretrained memory bank, there is a comparatively large performance improvement. This proves that with the memory bank pretrained on radiology reports, some prior knowledge (or language-derived knowledge) such as the topics in the report can be learned (the t-SNE visualization of the memory bank also demonstrate this, see Appendix), and the language-derived prior knowledge makes contribution to the cross modal understanding. What's more, Figure \ref{figure:4} shows the BLEU-4 and ROUGE-L scores during training, we can see that with the help of pretrained memory, it's easier for the model to achieve better performance with less training epochs and the training process of the model is relatively more stable. After adding the cross modal semantic alignment module (Trial 2, 5), we can see that all the scores of the metrics increase substantially. This shows that the constraints put on the memory retrieval process indeed helps the model to retrieve prior knowledge more effectively and perform better in the generation task. It is also noteworthy that the improvement made by adding learnable prompts in the decoder is comparatively less than other proposed modules in our method. The reason for this may be the randomness of the prompts initialization.

Table \ref{table:3} assesses the impact of memory bank capacity (i.e., $N^m$) tested on different dataset. We train the models with the same model structure and use the same hyperparameters (according to section IV.B) except for $N^m$. Models with larger capacities seem to have better performance. The performance scores peak when the capacity of the memory bank is 2048, and they don't increase when we use larger memory bank capacities. We also notice that the models trained on IU-Xray dataset have more severe performance decrease compared with those trained on MIMIC-CXR dataset when $N^m$ is larger than 2048. The reason for this may be that MIMIC-CXR dataset has more data than the IU-Xray dataset, so that it's harder to train the memory bank with larger capacity efficiently. Table \ref{table:4} evaluates the number of selected top $k$ elements in the memory bank. When $k \le 32$, the model performance increases when $k$ becomes bigger. However, if $k$ is larger than 32, the score decreases. Actually, when we use too much retrieved memory for each representation (when $k$ is big), more parameters in the memory bank will be updated, and this may be deleterious especially when we want to ensure the diversity of information stored in the memory bank (since the parameters update too frequently).

Figure \ref{figure:5} is the visualization of memory retrieval process, where the red arrows and the black dotted arrows denote the retrieval process of MCSAM and BASE, respectively. We can learn that the representations of different modalities but have the same semantic information such as "nodule" retrieve the same part of the memory bank, while the BASE model shows lower consistency. Figure \ref{figure:6} shows the length distribution of reports generated by BASE model and our MCSAM. We can see the great improvement of the ability to generate long reports after using our method.

\subsection{Further Discussion}

\begin{table*}[h]
  \caption{Further discussion of the cross modal semantic alignment module, where \textbf{SAM-dual} is the dual encoder form of SAM, \textbf{-w/o ADD} means whether the semantic visual feature embeddings provided by SAM is added to the visual encoder or not and \textbf{PreMem} denotes the use of the memory initialization trick. }
  \centering
  \label{table:5}
  \begin{tabular}{ccccccc}
    \toprule
    Method & BLEU-1 & BLEU-2 & BLEU-3 & BLEU-4 & METEOR & ROUGE-L \\
    \midrule
    BASE + PreMem + SAM-dual-oADD & 0.371 & 0.225 & 0.148 & 0.105 & 0.144 & 0.278 \\
    BASE + PreMem + SAM-dual-wADD & 0.373 & 0.227 & 0.150 & 0.106 & 0.146 & 0.277\\
    BASE + PreMem + SAM & \textbf{0.376} & \textbf{0.229} & \textbf{0.152} & \textbf{0.108} & \textbf{0.147} & \textbf{0.279} \\
    \bottomrule
  \end{tabular}
\end{table*}

\begin{figure*}
  \centering
  \includegraphics[width=0.9\linewidth]{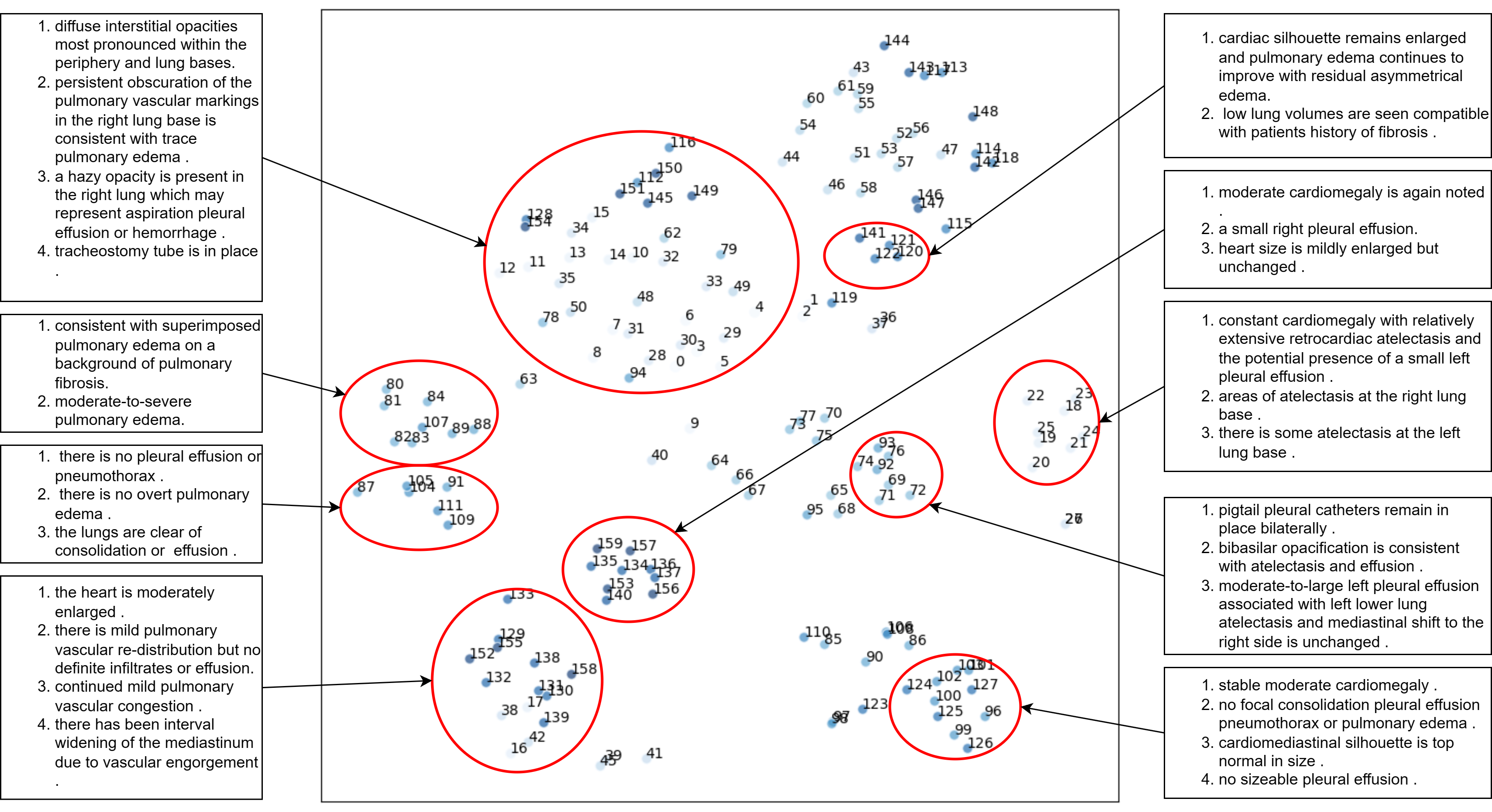}
  \caption{t-SNE visualization of the semantic visual feature embeddings. The red circles denote the clusters of the embeddings, the sentences are the main information of the corresponding reports of the radiology image and include the semantic meanings of the embeddings (e.g., different diseases or abnormalities ) }
  \centering
  \label{figure:7}
\end{figure*}

\begin{table}
  \caption{The Effect of batch size (the experiments are conducted on MIMIC-CXR dataset).}
  \label{table:6}
  \centering
  \begin{tabular}{cccc}
    \toprule
      batch size & BLEU-4 & METEOR & ROUGE-L \\
    \midrule
    8 & 0.102 & 0.140 & 0.276\\
    16 & \textbf{0.109} & 0.149 & \textbf{0.284}\\
    32 & 0.108 & \textbf{0.150} & 0.282\\
    64 & 0.107 & 0.148 & 0.280\\
    \bottomrule
  \end{tabular}
\end{table}

\textbf{The Effect of SAM.} SAM receives prior knowledge retrieved from different modalities and outputs the semantic visual feature embeddings. The experiments that we conduct are listed in Table \ref{table:5} to validate the effectiveness of SAM.

In Table \ref{table:5}, we learn that with the semantic visual feature embeddings added to the visual encoder, the model performance improves. Since acting as the topic of radiology images, the semantic visual feature embeddings make contributions to the report generation process. Besides, additional constraints are made by this simple operator since the semantic visual feature embeddings are derived from the memory bank. Thus, it is beneficial for the model to learn a better memory bank.

What's more, in Figure \ref{figure:7}, we show the t-SNE visualizations of the semantic visual feature embeddings and relate them to the main information of their corresponding reports. The red circles denote the clusters of the embeddings, the numbers represent the semantic visual feature embeddings of different radiology images and the sentences are the main information of the corresponding reports of the radiology image and include the semantic meanings of the embeddings (e.g., different diseases or abnormalities). We note that the semantic visual feature embeddings have clear clustering structure and embeddings having different semantic meanings are well separated from each other. This proves that the SAM can generate semantic feature embeddings with rich semantic meanings and help generate accurate radiology reports with high factual completeness.

In addition, we can also learn that although the siamese structure has fewer parameters than the dual encoder structure, it performs better. It's quite interesting that the siamese structure is widely used in some single modality contrastive learning approaches, such as sentence embedding models. Here, the cross modality task produces natural augmentations since the representations of different modalities coming from the same (image, text) pair can be used as positive pairs and those coming from different (image, text) pairs can be used as negative pairs.

In our approach, We use a contrastive loss to align the semantic information of different modalities. In many contrastive learning methods\cite{chen2020simple, he2020momentum, chen2021exploring}, the model performance is highly related to batch size. Table \ref{table:6} shows the batch size we choose to train the model, we can see that when we use a relatively larger batch size, such as 16 and 32, better performance is achieved (compared to the experiment using batch size 8), which is consistent with previous work (However, due to limited GPU memory, we cannot test our model on larger batch sizes such as 128.).

\textbf{The Effect of Different Alignment Methods.} Except for different applications of SAM, we also conduct some experiments to validate the effectiveness of two kinds of alignment approaches which are high-granularity alignment (HGA) and semantic alignment (i.e., SAM). As for the high-granularity alignment, when a single (image, text) pair is given,  we use the prior knowledge $P^S\in\mathbb{R}^{N_l\times D}$ and $P^V\in\mathbb{R}^{N_I\times D}$ retrieved from the memory bank and use the cross attention mechanism to generate two cross modal prior embeddings:

\begin{equation}
\begin{split}
        C^{MS} = Softmax(\frac{P^SP^V}{\sqrt{D}})P^V, \\
        C^{MV}=Softmax(\frac{P^VP^S}{\sqrt{D}})P^S,
\end{split}
\end{equation}
where  $C^{MS}$  and $C^{MV}$ are cross modal textual and visual prior embeddings, respectively. 

Since $C^{MS}$ is actually constructed by the visual prior knowledge and $C^{MV}$ is constructed by textual prior knowledge, we then formulate the contrastive losses $L_{txt}$, $L_{img}$ between the cross modal prior embeddings and their corresponding prior knowledge:
\begin{equation}
\begin{split}
    L_{txt}=-\frac{1}{N_{l}}(\sum_{i=1}^{N_{l}} log(\frac{exp \ cos(P^S_i, C^{MS}_i)/ \tau}{\sum_{j=1}^{N_{l}}exp\  cos(P^S_i, C^{MS}_j)/ \tau}) \\
    + \sum_{j=1}^{N_{l}}log(\frac{exp \ cos(P^S_j, C^{MS}_j)/ \tau}{\sum_{i=1}^{N_{l}}exp  \ cos(P^S_i, C^{MS}_j)/ \tau})),
\end{split}
\end{equation}

\begin{equation}
\begin{split}
    L_{img}=-\frac{1}{N_{I}}(\sum_{i=1}^{N_{I}} log(\frac{exp \ cos(P^V_i, C^{MV}_i)/ \tau}{\sum_{j=1}^{N_{l}}exp\  cos(P^V_i, C^{MV}_j)/ \tau})\\
    +\sum_{j=1}^{N_{I}}log(\frac{exp \ cos(P^V_j, C^{MV}_j)/ \tau}{\sum_{i=1}^{N_{I}}exp  \ cos(P^V_i, C^{MV}_j)/ \tau})).
\end{split}
\end{equation}

Then, the high-granularity contrastive loss of a single (image, text) pair in a batch is the sum of $L_{txt}$ and $L_{img}$ :
\begin{equation}
    L_{granu\_contrast}=\lambda_1 L_{txt}+\lambda_2 L_{img},
\end{equation}
where $\lambda_1$ and $\lambda_2$ are two hyperparameters which control the contribution of contrastive losses for different modalities.

\begin{table}
  \caption{Further discussion of different alignment methods. \textbf{BASE (+PreMem)} is the BASE model with initialized memory bank.}
  \label{table:7}
  \centering
  \begin{tabular}{ccccc}
    \toprule
    Metrics  &  BASE (+PreMem)  & +HGA & +SAM \\
    \midrule
    BLEU-1 & 0.356 & 0.361 & \textbf{0.376}\\
    BLEU-2 & 0.218 & 0.219 & \textbf{0.229}\\
    BLEU-3 & 0.144 & 0.145 & \textbf{0.152}\\
    BLEU-4 & 0.102 & 0.102 & \textbf{0.108}\\
    METEOR & 0.143 & 0.141 & \textbf{0.147}\\
    ROUGE-L & 0.276 & 0.275 & \textbf{0.279}\\
    \bottomrule
  \end{tabular}
\end{table}

\begin{figure}[h]
  \centering
  \includegraphics[width=\linewidth]{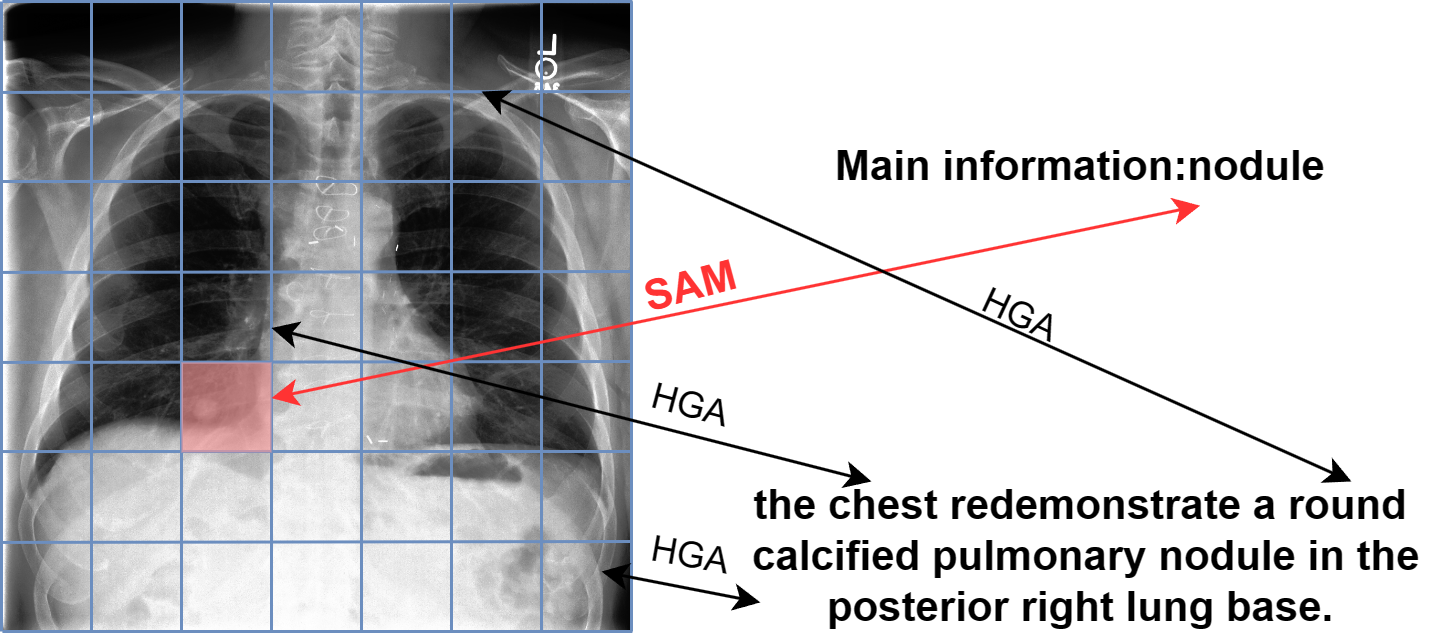}
  \caption{Illustration of SAM and HGA. HGA tries to align all the parts of the radiology images and each word in their corresponding reports while SAM only aligns the most important information in different modalities.}
  \label{figure:8}
\end{figure}

We compare the performance of HGA and SAM, which is showed in table \ref{table:7}. It can be seen that SAM surpasses the HGA with a large margin and HGA has worse performance than BASE model in some metrics. Practically, HGA tries to align all the parts of the radiology images and each word in their corresponding reports. However, abnormalities and key words only exist in small parts of radiology images and reports, and most of the representations which are aligned by HGA have no complete semantic meanings such as prepositions in the reports and the corners of the images. Thus, disturbance is made by HGA and this makes the training process unstable, while SAM uses the siamese Bert to get main information of different modalities and makes more interpretable alignment. The illustration of SAM and HGA is showed in Figure \ref{figure:8}.

\textbf{The Effect of Learnable Prompts.}
\begin{table}
  \caption{Further discussion of Learnable Prompts. \textbf{BASE (+PreMem +SAM)} is the BASE model with initialized memory bank and cross modal semantic alignment module. }
  \label{table:8}
  \centering
  \begin{tabular}{cccc}
    \toprule
    Metrics  &  BASE (+PreMem +SAM) & +parameter & +wm\\
    \midrule
    BLEU-1 & 0.376  &  0.367 & \textbf{0.379}\\
    BLEU-2 & 0.229  &  0.225 & \textbf{0.230}\\
    BLEU-3 & 0.152  &  0.151 & \textbf{0.153}\\
    BLEU-4 & 0.108  &  0.107 & \textbf{0.109}\\
    METEOR & 0.147  &  0.144 & \textbf{0.149}\\
    ROUGE-L & 0.279  &  0.279 & \textbf{0.284}\\
    \bottomrule
  \end{tabular}
\end{table}
Prompts are usually used in large language models (LLM) such as GPT to generate better sentences. Some previous works \cite{bulatov2022recurrent,li2023efficient} also explored the combination of additional learnable parameters and transformer models. We add the learnable prompts to learn some states or additional information when generating reports. For better understanding the effect of learnable prompts, in Table \ref{table:8}, we compare our method with directly increasing the dimension of feed forward layers (linear transformations) in the decoder (i.e., +parameter). We can observe that the learnable prompts actually perform better. This may suggest that learnable prompts are more efficient for training. What's more, although our motivation is to develop a model trained with no human annotations, more potential of learnable prompts can be probably explored by harnessing them for diagnose diseases or predicting the bounding boxes for the region with abnormalities when human annotations are available.

\section{Conclusion}
In this paper, we proposed MCSAM to generate accurate and fluent radiology reports according to radiology images. To learn disease-related representations and prior knowledge for different modalities, a memory bank initialized by learning the topics of the reports is proposed. Then, a memory retrieval process based on cross attention mechanism together with a cross modal semantic alignment module (SAM) are used to retrieve high consistency cross modal prior knowledge and make fine-grained feature consolidation. SAM also provides semantic visual feature embeddings which can be used in the decoder to benefit report generation. What's more, some randomly initialized parameters which act as prompts are added to the decoder to memorize additional information which is important to generate fluent reports. Experimental results show the effectiveness of our proposed model which outperforms many SOTA methods. Further analysis demonstrate the interpretability of our method.
\\
\textbf{Future Approach.} For better leveraging existed human annotations (if available), more efforts can be done for the improvement of our method: 1) Use the semantic feature embeddings generated by cross modal semantic alignment module for disease classification, which may help generate embeddings with more information and thus facilitating cross modal alignment; 2) With human annotations, we can fully harness the learnable prompts to diagnose diseases and even predict the bounding boxes for the region with abnormalities. This will not only benefit model performance, but also produce additional insights to help radiologists diagnose diseases.

\section*{References}
\bibliographystyle{ieeetr}
\bibliography{ref}



\end{document}